\newif\ifplainarticle
\ifdefined\PLAINARTICLE
  \plainarticletrue
\else
  \plainarticlefalse
\fi

\newif\ifanonsubmission
\ifdefined\ANONSUBMISSION
  \anonsubmissiontrue
\else
  \anonsubmissionfalse
\fi

\ifplainarticle
  \documentclass[10pt,twocolumn]{article}
  \usepackage[a4paper,margin=0.72in,columnsep=0.24in]{geometry}
  \usepackage[numbers,sort&compress]{natbib}
\else
  \ifanonsubmission
    \documentclass[sigplan,10pt,anonymous,review]{acmart}
  \else
    \documentclass[sigplan,screen]{acmart}
  \fi
  \setcopyright{none}
  \renewcommand\footnotetextcopyrightpermission[1]{}
\fi

\usepackage[T1]{fontenc}
\usepackage[utf8]{inputenc}
\usepackage{booktabs}
\usepackage{tabularx}
\usepackage{array}
\usepackage{xspace}
\ifplainarticle
  \usepackage{amsmath,amssymb}
\else
  \usepackage{amsmath}
\fi
\ifplainarticle
  \usepackage[hidelinks]{hyperref}
\fi

\newcommand{\lean}{Lean~4\xspace}
\newcommand{\op}{\mathbin{\diamond}}
\newcommand{\Fin}{\mathop{\mathrm{Fin}}}
\newcommand{\Nat}{\mathbb{N}}
\newcommand{\code}[1]{\texttt{#1}}
\newcommand{\prov}[2]{\begingroup\scriptsize\raggedright\nolinkurl{#1}\\[-1pt]\texttt{#2}\par\endgroup}
\newcommand{\pending}{\mbox{\textbf{PENDING}}}
\newcolumntype{Y}{>{\raggedright\arraybackslash}X}
\newcolumntype{R}{>{\raggedleft\arraybackslash}p{0.65in}}

\title{A Certificate-Producing Cascade for Equational Implication:\
The SAIR EQT2 Stage 2 Solver}

\ifplainarticle
  \ifanonsubmission
    \author{Anonymous submission}
  \else
    \author{Haobo Ma\\ChronoAI Pte.\ Ltd., Singapore
      \and Wenlin Zhang\\National University of Singapore
      \and Manuel Israel C\'azares\\Bytepro AI, Mazatl\'an, Mexico}
  \fi
  \date{}
\else
  \author{Haobo Ma}
  \affiliation{\institution{ChronoAI Pte.\ Ltd.}\country{Singapore}}
  \author{Wenlin Zhang}
  \affiliation{\institution{National University of Singapore}\country{Singapore}}
  \author{Manuel Israel C\'azares}
  \affiliation{\institution{Bytepro AI}\city{Mazatl\'an}\country{Mexico}}
\fi

\begin{document}
\ifplainarticle
  \maketitle
\begin{abstract}
The SAIR Mathematics Distillation Challenge on Equational Theories asks a
solver to classify whether one magma identity implies another and, for either
verdict, to return a certificate accepted by a deterministic \lean judge.
We present a single-file solver organized as a cheapest-first cascade.  Its
false branch combines coefficient tests over structured algebra families,
bounded finite-model search, an explicit central-groupoid witness, and several
infinite-carrier witnesses.  Its true branch is a proof-producing ordered unit
superposition procedure with Knuth--Bendix ordering, bidirectional
demodulation, indexing, memoised substitution, and anytime size deepening.
Search results remain outside the trusted base: successful derivations are
replayed as small \lean terms, and countermodels are rechecked by the
competition judge.

The frozen solver is a 189,504-byte Python file with SHA-256
\texttt{f2392533c9f4c03b...}.  In local runs through official judge revision
\texttt{2848228}, it produced accepted certificates for all 1,889 rows of the
six public sets with no language-model calls.  Separate measurements recorded
full agreement on the 800 published Stage 1 evaluation-distribution problems,
100 accepted rows in the canonical Marathon manifest without tokens, and 200
accepted rows in the hosted playground.  These are regression and playground
measurements, not a leaderboard result and not evidence about a hidden set.
All quantitative claims are tied to immutable result ledgers; the paper makes
no completeness or comparative-superiority claim.
\end{abstract}
\else
\begin{abstract}
The SAIR Mathematics Distillation Challenge on Equational Theories asks a
solver to classify whether one magma identity implies another and, for either
verdict, to return a certificate accepted by a deterministic \lean judge.
We present a single-file solver organized as a cheapest-first cascade.  Its
false branch combines coefficient tests over structured algebra families,
bounded finite-model search, an explicit central-groupoid witness, and several
infinite-carrier witnesses.  Its true branch is a proof-producing ordered unit
superposition procedure with Knuth--Bendix ordering, bidirectional
demodulation, indexing, memoised substitution, and anytime size deepening.
Search results remain outside the trusted base: successful derivations are
replayed as small \lean terms, and countermodels are rechecked by the
competition judge.

The frozen solver is a 189,504-byte Python file with SHA-256
\texttt{f2392533c9f4c03b...}.  In local runs through official judge revision
\texttt{2848228}, it produced accepted certificates for all 1,889 rows of the
six public sets with no language-model calls.  Separate measurements recorded
full agreement on the 800 published Stage 1 evaluation-distribution problems,
100 accepted rows in the canonical Marathon manifest without tokens, and 200
accepted rows in the hosted playground.  These are regression and playground
measurements, not a leaderboard result and not evidence about a hidden set.
All quantitative claims are tied to immutable result ledgers; the paper makes
no completeness or comparative-superiority claim.
\end{abstract}
  \maketitle
\fi

\ifplainarticle\else
\fi

\section{Introduction}

An equational theory of magmas has one binary operation and no assumed
associativity, identity, or cancellation law.  Given equations
$E_1$ and $E_2$, the implication problem asks whether every magma satisfying
$E_1$ also satisfies $E_2$.  The SAIR Mathematics Distillation Challenge,
organized by Damek Davis and Terence Tao with the SAIR Foundation, turns that
question into a certificate-production task grounded in the Equational
Theories Project~\cite{competition,etp}.  A true answer must contain a \lean
proof of the universal implication.  A false answer must exhibit a magma in
which the hypothesis holds and the conclusion fails.  The same deterministic
judge checks both forms.  An accepted certificate solves the row; a plausible
label, proof sketch, or partly checked derivation receives no credit
\cite{competition,competitionRules}.

This contract changes the design objective.  A classifier may be useful while
remaining difficult to audit, whereas a certificate producer must preserve
enough structure to reconstruct a kernel-checkable witness.  At the same time,
the kernel boundary permits aggressive untrusted search.  A defect in our term
ordering, unifier, model finder, or certificate printer may increase resource
use or prevent a solution from being found, but it cannot cause an invalid
certificate to be accepted.
The competition proxy records success only after the official judge has
compiled the submitted term and checked its dependency policy.

Our system implements a deterministic-first cascade in one Python file.  The
early stages recognize direct implications and search inexpensive countermodel
families.  A bounded irregular model finder and deeper structured families
handle false residuals.  An ordered unit-superposition engine handles true
residuals and records a replayable derivation rather than only a refutation
bit.  The same source dispatches to the Solo protocol, in which one process
handles one problem, or the Marathon protocol, in which one process shares a
global budget among a manifest of problems.  Every successful path ends in a
\lean certificate.

The central empirical result is deliberately narrower than a completeness
claim.  The cascade covers every row in the measured public,
distribution-drill, and hosted-playground corpora within the recorded budgets.
Neither the countermodel families nor the superposition implementation is
proved complete, and the deeper stages stop at wall-clock limits.  We therefore
do not infer performance on the held-back evaluation set.  The result is a
system description and an auditable regression report, not a theorem about all
single-identity implications.

This paper makes four contributions.
\begin{itemize}
  \item We present a certificate-producing cascade that integrates
  complementary algebraic model families and proof-producing superposition
  behind one competition interface.
  \item We describe certificate encodings that satisfy the judge's dependency
  policy, including encodings for finite carriers beyond the direct table
  helper's single-digit range and core-only proofs over infinite carriers.
  \item We analyze two consequential implementation failures: retention of
  tautologies in the active superposition set and inconsistent handling of the
  two accepted spellings of the magma operation.  End-to-end certificate runs,
  rather than label accuracy alone, exposed both failures.
  \item We provide a hash-bound claim ledger that connects the solver bytes,
  judge revision, result files, and every evaluation-table row to SHA-256
  identifiers.
\end{itemize}

The rest of the paper follows the execution path.  Section~\ref{sec:arch}
gives the single-file architecture.  Sections~\ref{sec:false} and
\ref{sec:true} describe the countermodel and proof branches.
Section~\ref{sec:hosted} covers deployment constraints, and
Section~\ref{sec:evaluation} reports measurements.  We then state the
reproducibility boundary and place the system relative to equational-theory,
superposition, and proof-assistant work.

\section{Single-File Solver Architecture}
\label{sec:arch}

\subsection{Contract and trusted boundary}

The submission contract requires one \code{solver.py} file no larger than the
competition limit.  Solver v2.5 is 189,504 bytes, and therefore remains below
that limit while containing all search code and fixed data.  In Solo mode the
proxy sends one JSON problem and its budget to a fresh process; the solver sends
judge or language-model requests back over line-delimited JSON.  In Marathon
mode the runner supplies a JSONL manifest and an append-only output path.  The
same source supports both modes, selected by the runner-owned Marathon
environment variable~\cite{competitionRules}.

The trusted computing base for a verdict is not the Python implementation.
The judge generates the problem statement and the verdict-specific \code{Goal}
type.  A true goal has the form
\[
  \forall (G : \mathsf{Type})\,[\mathsf{Magma}\;G],\quad E_1(G)\to E_2(G),
\]
whereas a false goal existentially quantifies a carrier and magma for which
$E_1$ holds and $E_2$ does not.  The solver supplies a definition named
\code{submission}; \lean elaboration, kernel checking, and the dependency
inspector determine acceptance.  Search is therefore an untrusted producer of
candidate proof objects.

The interface creates three distinct obligations.  First, a candidate must
have the right polarity: a universal implication for a true verdict or an
existential separating structure for a false verdict.  Second, the term must
inhabit the judge-generated goal rather than a solver-reconstructed copy that
might differ in variable order or parsing.  Third, all transitive dependencies
must satisfy the proof policy.  The judge enforces the first two by type
checking \code{submission : Goal}; its inspection pass enforces the third.
The emitted source never uses admitted terms.  This separation is why a
successful Python-side check is called a candidate throughout the
implementation, while ``accepted'' is reserved for the judge status.

The polarity boundary also determines failure recovery.  The solver does not
cache an unverified Boolean verdict and ask later stages merely to justify it.  Each
stage returns a self-contained candidate of one polarity, submits that
candidate, and continues if checking fails.  For example, an arithmetic
countermodel whose certificate exceeds an instance budget does not bias the
proof branch toward false; it is discarded as an unavailable witness.  A
superposition trace that cannot be replayed is likewise not evidence for a
true output.  This discipline prevents a search heuristic from escaping into
the externally visible result.

At both protocol entry points we normalize the input operation symbol.  The
official problem format permits either an asterisk or a diamond, but the
solver's term parser has one internal token.  Mapping the asterisk to
$\op$ before parsing preserves the semantics and is operationally necessary;
Section~\ref{sec:encoding} records the failure that revealed it.

\subsection{Cheapest-first cascade}

The Solo cascade is ordered by expected cost and by which branch benefits from
remaining time.

\paragraph{Stage A: shallow countermodels.}
The solver first tries tiny brute-force tables, scalar linear and affine
operations, vector-linear families, low-degree polynomial operations, and
recognized infinite-carrier Austin models~\cite{austin}.  The expanded structured families
add about 0.8 seconds on a true input in the recorded development measurement.
No irregular finite-model search is performed yet.

\paragraph{Stage B: direct true certificates.}
Syntactic singleton collapse and substitution-instance checks produce direct
proofs.  A specialized singleton-forced prover then receives an eight-second
cap, although applicable cases normally close during its initial search.  The
general superposition engine follows with a quick budget
\(
\min(6,\max(1,T/600))
\)
seconds for a process allowance $T$; this is six seconds under the recorded
Solo allowance.  Most provable public instances terminate within this cap.

\paragraph{Stage C: bounded irregular models.}
A finite-model search over carriers four through ten receives eight seconds.
It propagates the universal instances of the hypothesis and Latin-square
constraints rather than enumerating complete tables.

\paragraph{Stage D: deep false search.}
Only residuals reach heavier polynomial and vector families, capped at sixty
seconds, followed by another finite-model pass capped at one hundred twenty
seconds.  Deep false search precedes the deepest proof pass so that a difficult
false row does not consume most of its allowance in proof saturation.

\paragraph{Stage E: anytime proof search.}
The ordered-superposition engine, in the tradition of completion-based
equational provers~\cite{knuthBendix,bachmairGanzinger}, receives
\(
\min(600,\max(20,T/6))
\)
seconds, which is six hundred seconds under the Solo allowance.  The engine
repeatedly raises term-size and variable caps.  If no direct refutation is returned, a
small pool of derived equalities can be emitted as exact lemmas followed by a
kernel-checked closing tactic.  A language-model feedback loop follows
the deterministic cascade, but none of the accepted rows in the archived
regression invoked it.

Every stage independently verifies its candidate before the solver stops.  A
rejected certificate proceeds to the next strategy; a successful search
without a valid emitted term is not counted as a solution.  This control flow
distinguishes search coverage from certificate robustness.

\subsection{Representation and dispatch}

Both branches share a small non-associative term representation.  Parsing
produces a binary tree with variables at leaves; no reassociation or
commutative normalization is performed.  Alpha-renaming maps variables to a
canonical local order for cache keys and family matching, but emitted proofs
retain a map back to the judge's binder order.  Operation count, variable
support, depth, and tree size are computed once and reused by scheduling,
countermodel probes, and proof search.

Dispatch relies on positive recognizers.  A direct proof stage
returns only after constructing a substitution or equality chain.  An Austin
or central-groupoid stage returns only when the hypothesis matches its stored
schema and a concrete evaluation separates the goal.  A family that is merely
likely from equation identifiers is not selected: identifiers are useful for
diagnostics, but the solver matches parsed laws.  This property supports
independent replay and handles inputs whose labels or variable names differ
from those in the public files.

Candidate validation is stratified.  Low-cost Python evaluation removes
malformed tables, wrong witnesses, and printer bugs before a judge call.  Judge feedback
then closes the semantic and dependency obligations.  The development harness
records both outcomes: a ``found'' countermodel rejected by \lean indicates an
emitter defect or an encoding mismatch, whereas the absence of a Python
witness within a deadline indicates a coverage limitation.  This distinction informed the large-carrier encoding correction in
Section~\ref{sec:allowlist}.

\subsection{Solo and Marathon}

Solo gives each row a fresh process and a fixed allowance.  Marathon instead
requires triage across a whole manifest.  The batch path stably sorts problems
by a structural cost derived from operation occurrences, variable count, and
equation length.  It then runs two passes.  The first gives every row the
shallow stages before any residual receives deep work.  The second revisits
only residuals.  A fair cap is recomputed from remaining wall-clock time
divided by remaining rows and clamped to the documented per-row interval.  At most fifty-five
percent of a residual's current slice goes to the false branch; the proof
branch receives the rest.

Marathon outputs constitute durable state.  Each answer is one JSON line, flushed and
synchronized before the next row.  If the runner terminates the process, all
complete earlier lines remain scoreable.  A process-level timer bounds legacy
search loops that predate the batch driver.  The batch path issues no speculative
language-model calls and leaves a residual unanswered rather than consuming
tokens without producing a certificate.

\subsection{Cascade invariants and failure isolation}

Stage ordering does not assert that either polarity is more likely.  Instead,
it minimizes expected certificate cost while reserving time for both
branches.  A structured false witness is attractive early because checking a
closed operation is usually cheaper than saturation.  Direct true patterns are
equally attractive because they avoid both table construction and general proof
search.  Measurements showed that most provable rows close near the start of
saturation, which motivates placing the quick G3 pass before irregular model
search.  Deep false search precedes deep G3 only after every low-cost recognizer
has failed.

The implementation maintains a cascade invariant: before entering a stage,
the solver has no accepted certificate, and every earlier candidate has either
been absent, failed local construction, or been rejected by the judge.  It does
not maintain the stronger and generally invalid invariant that earlier stages
have ruled out their polarity.  Failure to find a finite model does not support
truth; failure to derive the goal does not support falsity.  Each stage can
therefore be analyzed and tested as a partial certificate producer, and the
composition is sound whenever the judge is sound, regardless of the producers'
coverage.

Exceptions are contained at strategy boundaries in the Marathon path so that a
malformed residual cannot discard previously completed rows.  Solo has a fresh
process per row and can rely on the runner to record a crash.  Within a
strategy, deadlines are checked at natural allocation points: carrier changes,
model-search decisions, deepening rounds, and given-clause iterations.  The
Marathon alarm provides a final outer bound for code paths whose internal
checks are too sparse.  Timeout returns ``no candidate'' rather than a verdict.

Caches follow the same lifetime discipline.  Parser and static-law data live
for the process.  A superposition run owns its unifier, normalization, term,
and clause caches, which are cleared before another problem.  Marathon could
in principle share more proof state across rows, but the current implementation
does not rely on cross-problem lemmas.  This keeps Solo and Marathon certificate
semantics aligned and limits memory growth under a long manifest.

The fallback model follows the complete deterministic cascade in Solo and is
not invoked in the archived accepted rows.  Its proposals still pass through
the same printer and judge, so adding the fallback does not enlarge the trusted
base.  In Marathon it is disabled to preserve the global token budget; the
result is a partial manifest containing only verified deterministic answers.
These are scheduling choices, not claims about the intrinsic capabilities of
the configured model.

\section{Countermodel Construction}
\label{sec:false}

For a false implication the solver must find an operation satisfying all
assignments of the hypothesis and at least one assignment falsifying the
goal.  It evaluates candidate families in Python, retains a concrete witness,
and emits a certificate in which the judge repeats the decisive checks.

\subsection{Coefficient matching over structured families}

Represent a magma term by the formal coefficients induced by a candidate
operation.  For
\[
  x \op y = ax+by \pmod n,
\]
each leaf contributes a coefficient obtained by multiplying edge labels along
its path.  Equality of two terms for all assignments is therefore equality of
their coefficient vectors modulo $n$.  This gives an exact, inexpensive test
for whether the hypothesis is an identity of the candidate magma.  The same
representation quickly locates an assignment on which the goal coefficients
differ.  The scan includes composite moduli and its shallow range extends to
the largest carrier admitted by the certificate-instance guard.

Affine operations add a constant term, and matrix-linear operations replace
the scalar coefficients by matrices over small finite fields.  The latter
searches include two-dimensional vectors over the three-element field and
three-dimensional vectors over the two-element field.  Polynomial candidates
are evaluated on full assignments when the carrier and hypothesis arity make
that feasible.  The deep pass adds a quadratic grid, a two-dimensional family
over the five-element field, and sampled polynomials of degree at most three.
An instance-count guard prevents a mathematically compact operation from
producing a certificate whose exhaustive hypothesis check exceeds the judge
budget.

These families provide both search efficiency and structured certificates.
They find many witnesses faster than a generic table search while preserving
semantic structure in the emitted certificate.  For example, a linear
countermodel can be represented by its arithmetic operation rather than by a
large opaque table.  The judge still evaluates the hypothesis and the failing
goal instance, so coefficient matching remains an untrusted search
optimization rather than a proof rule.

Coefficient evaluation is implemented recursively.  A variable leaf maps to
its basis vector.  At an internal node, the coefficient vector is
$a\vec{u}+b\vec{v}$ for child vectors $\vec{u}$ and $\vec{v}$; affine families
carry one additional constant coordinate.  Thus the universal hypothesis test
does not enumerate assignments.  If the two sides have different vectors,
the candidate is immediately rejected as a model of the hypothesis.  If the
hypothesis vectors agree and the goal vectors differ, a separating assignment
can be sought in the small module.  Full evaluation is still applied before
emission, which protects the certificate path from an error in this symbolic
shortcut.

Vector-linear families use exactly the same recursion with matrices acting on
coordinates.  They are useful because scalar coefficient coincidence may
hold in every small modulus even when non-commuting linear actions distinguish
the goal.  Polynomial operations abandon the linear invariant and instead use
compiled term evaluators over bounded carriers.  The family ordering therefore
progresses from symbolic coefficient equality to increasingly expensive full
evaluation, while presenting a uniform result to the certificate printer: a
carrier, a total binary operation, and a goal witness.

\subsection{Irregular finite models}

Some countermodels have no low-degree algebraic description.  The finite-model
stage treats the Cayley table as a constraint problem.  For a proposed carrier
size, table cells begin unassigned.  Ground instances of the hypothesis watch
the cells on which their two terms depend.  Assigning a cell may simplify a
term, force another cell, or violate an instance.  The search uses a
minimum-remaining-values choice among table positions and abandons partial
tables as soon as a hypothesis instance becomes unequal.

Many identities imply that rows, columns, or both must be permutations.  A law in
which a variable occurs once on one side and appears under the operation in a
particular argument position can expose such a constraint.  The solver infers
the applicable Latin flags, rejects duplicate row or column values, and forces
naked singletons.  For one residual requiring an eight-element quasigroup,
this propagation changed the outcome from no witness within a
one-hundred-twenty-second attempt to a witness in 0.3 seconds.  The production
shallow search remains restricted to carriers four through ten and an eight-second cap; the deep pass revisits the
same carrier interval with a longer cap.

Before emission, the completed table is checked by a separate evaluator over
all hypothesis assignments and a stored goal-falsifying assignment.  The
\lean certificate then reconstructs the table operation and invokes the
judge's finite-decision tactic.  The duplicated checks reduce wasted judge
calls, while only the latter determines acceptance.

The partial-table solver maintains two invariants.  Every assigned cell agrees
with all hypothesis instances that have become ground, and every inferred
Latin constraint agrees with the assigned portion of its row or column.  A
watch list connects a cell to the term instances whose evaluation may advance
when that cell is filled.  Propagation reaches a fixed point before the next
decision.  At a complete table, the first invariant implies that exhaustive
reevaluation of the hypothesis should succeed; the separate final evaluator
checks this implication and also searches the goal assignments.  Search
backtracking may be unsound or incomplete as an algorithm without threatening
an accepted answer, because the table is subsequently evaluated from scratch
and then checked once more in \lean.

Carrier growth is likewise heuristic.  The shallow and deep passes enumerate
the documented carrier interval, but the solver does not infer that absence of
a model there implies truth.  It passes the row to the next proof or
model family.  This behavior implements the paper's no-completeness boundary.

\subsection{Central groupoids and the E168 residual family}

The evaluation-distribution drill identified a family of twelve residuals whose
hypothesis is the central-groupoid law.  Central groupoids have a well-known
combinatorial structure studied by Knuth~\cite{knuthCentral}.  The natural
constructions tried by the structured stages satisfy both the hypothesis and
the relevant E168 conclusions, so they do not separate these equations.
Increasing the generic finder budget did not reliably recover the missing witnesses.

The solver therefore includes one explicit non-natural central groupoid of
order nine.  The table was checked against each relevant hypothesis and goal,
then stored once as a fixed witness family.  When the hypothesis matches up
to variable renaming and a goal is separated by the table, the solver emits
the ordinary finite-table certificate.  This stage is not a classification of
central groupoids and makes no claim about minimal order.  It provides a
specific, auditable witness for a recurring algebraic family: the official
runner accepted the emitted certificates for the twelve drill residuals.

\subsection{Infinite carriers for Austin pairs}

Finite satisfiability can conceal false general implications.  Several pairs
catalogued in the Equational Theories Project are true over the finite magmas
searched by standard methods but false in general.  For recognized Austin-pair
hypotheses, the solver selects a fixed operation on $\Nat$, or its opposite,
and a small tuple falsifying the goal.  The certificate proves the hypothesis
through a fixed sequence of elementary lemmas and closes the negated goal with
the concrete tuple.

This is qualitatively different from finite decision.  There is no exhaustive
tactic over $\Nat$, so the complete mathematical argument must be expressed in
the certificate.  The original certificates imported a broad tactic module;
Section~\ref{sec:leanmigration} describes why those imports became an
operational problem and how positive-form replay reduced their compile time.

\subsection{Certificate shapes and the allow-list}
\label{sec:allowlist}

For carriers through ten, \code{finOpTable} embeds a JSON-like Cayley table and
\code{decideFin!} checks the finite existential goal.  The helper extracts
individual decimal characters, so a table entry with multiple digits is not a
faithful encoding.  Larger carrier tables therefore cannot use that direct
shape.

The alternative is a definition such as \code{submission.op} whose body
computes a linear, affine, polynomial, or packed-table lookup operation.  The
judge checks its allow-list on the direct constants of the submitted term;
helpers inside the \code{submission} namespace are unrestricted, which the
organizers document as intended for auxiliary definitions.  The final theorem
refers directly to the namespaced operation and core arithmetic.  This route
was accepted by the official judge for the recorded larger carrier sizes and
avoids relying on the table parser.

This distinction affects dependency enforcement but does not circumvent it.
The helper definition is part of the submitted source, is compiled by \lean, and is traversed by the
kernel when its theorem is checked.  The allow-list controls dependencies, not
the truth of definitions.  As specified by the documented proof interface,
submission-local auxiliary definitions provide the intended mechanism for
packaging operations and lemmas that are not global library dependencies.

\section{Proof-Producing Ordered Superposition}
\label{sec:true}

The true branch operates in unit equational logic.  It assumes one universally
quantified identity and attempts to derive the goal identity.  The engine is
inspired by ordered superposition and unfailing completion as implemented in
first-order provers, but is specialized to one binary symbol, unit clauses,
and the need to reconstruct a small positive \lean proof.

\subsection{Terms, ordering, and inference}

Terms are immutable trees over variables and the binary magma symbol.  A
Knuth--Bendix ordering with unit symbol weights determines maximal equation
sides and orients usable rewrite rules.  Superposition overlaps an oriented
side with a non-variable subterm of another maximal side, computes a most
general unifier, and constructs the resulting equality.  Ordering checks are
repeated after unification, because substitution can change comparisons.
Equations that cannot be oriented remain available for symmetric inferences.

The goal is represented internally as a negative unit clause $u\neq v$ over
fresh constants.  Active equalities can superpose into either side; ordinary
demodulation simplifies the clause.  If its sides unify, the negative clause
is refuted.  This internal negative representation is convenient for search,
but the emitted proof does not ask \lean to validate a refutation calculus.
It instead reconstructs a positive equality chain ending at the original
goal.

Forward demodulation normalizes each generated equality by current rules.
Backward demodulation revisits active equations when a new rule can simplify
them.  Symmetric alpha-normalized keys remove variants.  A given-clause loop
interleaves a weight heap with an age queue; later deepening rounds alter the
variable penalty and age ratio so that a proof blocked by one selection policy
is not permanently starved.

\subsection{The tautology-deletion failure}

An early version of the prover did not resolve one public residual,
\code{hard3\_0314}.  The official-runner attempt found no proof after 810 seconds
even though an external derivation suggested that the necessary clauses were within the
configured size bound.  Instrumentation indicated that throughput was not the
primary cause.  When demodulation transformed an equation into $s=s$, the
step constructor returned no proof node, and the caller retained the partially
rewritten input equation.  Backward simplification had the same behavior.

Consequently, many small instances of the hypothesis remained live.  Their
weights made them attractive to given-clause selection, so they repeatedly
displaced clauses on the useful derivation.  The correct simplification rule
is deletion: a demodulated tautology contributes nothing to saturation.  The
fix retires the active clause before testing the rewritten result and drops a
new clause whose sides become equal.

With that change, the prover found the residual proof in 0.6 seconds in its
direct measurement, and the official-runner path produced an accepted
certificate in 14.6 seconds of wall-clock time.  These measurements illustrate
the coupling between simplification and selection rather than establishing a
general speed ratio.  Retaining a tautology as a live clause changed the
effective search strategy enough to prevent the final public proof from being
found within the earlier attempt.

\subsection{Indexing and allocation control}

Demodulation candidates are retrieved through a discrimination tree keyed by
a term's linear skeleton.  The index returns rules whose left sides may match a
subterm; matching and the ordering check filter false positives.  Normal forms
are cached and versioned by the rewrite-rule set.  Term sizes, variable
multisets, preorder traversals, and eligible positions are cached on immutable
equations.

The unifier returns a triangular substitution rather than eagerly applying it
to every binding.  A memoised substitution routine resolves chains while
sharing unchanged subtrees.  Before constructing a superposition result, the
engine estimates instantiated size from static term sizes and per-variable
counts.  Overlaps that cannot fit the current cap are rejected before
allocating their terms.  Unifier results are cached by the renamed overlap
pair.  These changes preserve the inference trace at fixed bounds while
reducing the work per surviving clause.

The passive set stores recipes---source equation, target equation, direction,
and overlap position---rather than full instantiated proof objects.  The
selected recipe is materialized when it becomes given.  Under the hosted
memory limit, this representation permits a large passive frontier without
duplicating every term and ancestry chain.  Caches and negative-goal storage
are bounded and cleared between prover invocations.

\subsection{Anytime deepening}

A fixed size cap is sensitive to the selected bound: saturation at a small cap
can terminate with time remaining even when one slightly larger intermediate would finish the proof.
The deep pass therefore runs a schedule of increasing term-size and variable
caps.  If a round saturates, its unused time moves forward.  If the final
configured round saturates, caps continue to grow within the overall deadline.
The last strategies reduce the variable penalty and increase age pressure,
approximating a different prover configuration without maintaining another
implementation.

The procedure remains budget-bounded and machine-load sensitive.  It is not a
new completeness result for ordered superposition, nor a proof that the chosen
ordering is fair under every cap.  The evaluation in
Section~\ref{sec:evaluation} measures this implementation on fixed corpora.

\subsection{Replay into the \lean kernel}

Every retained equation stores a derivation recipe.  Replay recursively
reconstructs source equations, instantiates the universally quantified
hypothesis, and uses congruence to place an equality inside the overlap
context.  A superposition step becomes a \code{have} binding assembled from
\code{congrArg}, symmetry, and transitivity.  Goal simplifications become
another exact equality chain.  The final proof is a sequence of
\code{have}/\code{rw}/\code{exact}-style core terms in positive form.

Positive replay isolates search from trust.  The Python engine can orient an
equation incorrectly, use a stale cache, or reconstruct the wrong path; any of
these produces a term that fails to elaborate or has the wrong type.  The
judge then rejects it, and the solver may retry robust emission or continue to
another stage.  A prover defect can therefore cause a missed proof or a slower
run, but cannot cause a wrong acceptance unless there is also a defect below
the \lean kernel boundary.

Replay granularity and certificate size impose competing requirements.
Emitting every normalization micro-step is easy to audit but repeats contexts
and substitutions.  Compressing an entire derivation into a large tactic call
reduces source size but reintroduces a dependency on tactic behavior and
library imports.  The solver uses one named equality per selected inference,
with normalization represented as explicit transitivity where needed.  Shared
ancestors are emitted once in topological order.  This keeps proof text close
to the saturation trace while allowing \lean to elaborate each local equality
without reconstructing the global search.

Robust re-emission addresses elaboration rather than logical inference.  The
first printer uses direct exact terms.  If the judge rejects that source, the solver can wrap
individual equality steps in a small sequence of core closing alternatives and
resubmit the same derivation; it does not rerun superposition or change the
claimed theorem.  If search exhausts its deadline without a refutation, the
lemma-pool path emits a bounded selection of the smallest already proved
consequences.  Any final tactic must still close the judge goal from exact
lemmas.  Failure to do so leaves the row unanswered.

\section{Engineering for the Hosted Environment}
\label{sec:hosted}

\subsection{Core-only certificates}
\label{sec:leanmigration}

The public harness used \lean 4.30.0-rc2, while the announced hosted verifier
used \lean 4.32.0 with the corresponding Mathlib release.  Most emitted
certificates already imported only judge modules and core definitions.  The
exception was the infinite-carrier Austin family, whose tactic import made
compilation sensitive to the larger library environment.  In a compatibility
run, one such certificate required about 330 seconds of compilation.

Rather than increasing the solver-side timeout, we rewrote the Austin
certificate in positive form, replaced the broad tactic dependency with
explicit core lemmas, and re-emitted the witness.  Its \lean 4.32 compilation
then completed within 3.1 seconds.  A stratified corpus of 120 accepted certificates, covering every emission family and all fourteen former v1
residuals, compiled 119 of 120 under the exact announced toolchain in an
external harness with a 120-second per-phase limit; the one non-pass was the
pre-rewrite Austin witness itself, which passes under a 300-second per-phase
configuration matching the judge (phases of 184 and 145 seconds).  The corpus
therefore characterizes the migration-era certificate set rather than the
frozen artifact's current output: the frozen solver's own certificate for the
same problem imports only the judge module and is accepted in under five
seconds.  The judge support modules compiled unchanged.  That corpus run executed outside the repository
and is archived as an operator attestation
(\code{results/lean432\_corpus/}).  When the organizers subsequently upgraded
the judge to \lean 4.33.1 after the kernel-soundness review, the frozen
artifact was revalidated in full under the upgraded judge --- official harness,
all six public sets, the distribution drill, and the Marathon manifest --- with
the resulting ledgers committed and hash-bound
(\code{results/lean4331\_revalidation/}).  Neither measurement substitutes for
hosted scoring.

\subsection{Input encoding}
\label{sec:encoding}

The official judge normalizes the asterisk and diamond spellings when it builds
the \lean problem.  The runner, however, passes the original problem text to
the contestant process.  An earlier solver normalized only along one internal
path.  When the Stage 1 evaluation splits were supplied in their published
asterisk encoding, the solver crashed before certificate search and recorded
zero accepted rows out of 800.

Version v2.4 introduced, and v2.5 retains, normalization of both equations at both the Solo and Marathon intakes.
The same drill then recorded 800 accepted certificates with full agreement
against the published answers.  Thus, verified output does not address failures
in an unverified parser; normalization must occur at the protocol boundary,
before any dispatch or caching.

\subsection{Resource and filesystem constraints}

The hosted-style sandbox has a read-only submission mount, a bounded process
count, bounded memory, no direct network, and a small writable temporary
filesystem.  The solver consequently uses the Python standard library, keeps
all static data in the single source file, does not spawn external provers, and
does not assume repository access.  Bounded passive queues and caches keep the
superposition engine below the sandbox memory ceiling observed in long false
saturations.  The process-level timer also prevents a deeply nested legacy
loop from overrunning its Marathon slice.

Marathon's output file is the only durable mutable state supplied by the
runner.  Per-row flush and synchronization are therefore required for
correctness.  An initial local fixture run also showed that the official
checkout's default artifact directory could be unwritable in a managed
workspace; redirecting judge artifacts to temporary storage resolved the
environmental failure without changing judge sources.

\section{End-to-End Certificate Production}
\label{sec:certificates}

Search and replay together form a single transaction.  This section follows
representative false and true candidates
from a parsed equation pair to the theorem checked by the judge.  The examples
show certificate shape rather than a particular benchmark row; the concrete
generated sources remain in the archived result ledgers.

\subsection{Finite countermodel transaction}

Suppose coefficient matching or table search returns a carrier $\Fin n$, an
operation $m$, and an assignment $\rho$ that separates the goal.  Before
printing \lean, the solver performs three checks.  It confirms that every
table entry or arithmetic result belongs to the carrier.  It enumerates every
assignment required by the hypothesis and checks equality of the evaluated
sides.  Finally, it reevaluates the two goal terms at $\rho$ and checks that
their values differ.  A failure at any point discards the candidate without a
judge call.

For a small table the emitted proof has the following schematic form:
\begin{verbatim}
def submission : Goal := by
  let m : Magma (Fin n) :=
    { op := finOpTable "<table>" }
  refine <Fin n, m, ?_>
  decideFin!
\end{verbatim}
The angle-bracket line is printed with \lean's tuple notation in the actual
source.  The judge's decision tactic expands the finite universal hypothesis
and the existential goal witness under the supplied operation.  Thus the
stored Python assignment is useful for prechecking, but the accepted theorem
does not rely on the Python evaluator.

For a larger structured carrier, the transaction changes only in how $m$ is
defined.  A helper inside the \code{submission} namespace computes the
operation using \code{Nat} and \code{Fin} primitives.  The theorem refers to
that helper when constructing the magma, and finite decision closes the same
logical goal.  Packed-table witnesses use a natural-number payload and an
indexing expression in the helper body.  Arithmetic families print their
closed form directly.  Both routes keep multi-digit entries away from
\code{finOpTable} while leaving the operation transparent to the kernel.

The certificate has two independent failure modes.  The operation may not be a
countermodel, in which case finite decision fails with an ordinary proof
error.  Or the source may use a dependency outside the policy, in which case
inspection classifies it as incomplete.  Development probes must run through
the complete official policy: an earlier permissive harness omitted the
declaration restriction and made a rejected arithmetic shape appear usable.
Only the official runner result was retained as evidence.

\subsection{Infinite countermodel transaction}

An infinite witness cannot delegate universal checking to finite reflection.
The Austin path first matches the parsed hypothesis against a stored law up to
renaming.  It selects an operation $f : \Nat\to\Nat\to\Nat$, instantiates a
fixed proof that $f$ satisfies the law, and evaluates candidate tuples until
the goal sides differ.  Emission packages $\Nat$, the magma instance, the
hypothesis theorem, and a function that turns any assumed goal equality at the
chosen tuple into a contradiction.

The hypothesis theorem is the expensive part of the source.  Its core-only
version is a sequence of named local facts about the operation followed by the
target identity.  The goal refutation is intentionally concrete.  It applies
the universally quantified conclusion to the stored natural-number tuple and
uses computation to show that the resulting equality is impossible.  This
division makes the reusable part of the mathematical construction visible
while keeping each problem-specific separation small.

As with finite models, recognizing the hypothesis is not itself trusted.  A
renaming bug would instantiate a lemma at the wrong law, and \lean would reject
the resulting type.  A wrong tuple would fail to close the negation.  The
kernel therefore checks both the reusable model theorem and its application to
the current judge-generated equations.

\subsection{Superposition replay transaction}

A successful G3 search returns a refuted negative goal plus a directed acyclic
graph of selected inference recipes.  Replay begins from the nodes reachable
from the refutation and topologically orders them; irrelevant saturation
clauses are omitted.  Hypothesis leaves become applications of the input
identity with a substitution for its universally quantified variables.
Derived leaves created by symmetry or reflexivity become the corresponding
core constructors.

For an overlap, replay first obtains the equality used as a rewrite.  It lifts
that equality through the surrounding term context with \code{congrArg},
choosing symmetry if the search used the reverse direction.  It then composes
the lifted equality with the instantiated target equation by transitivity.
The generated source consequently resembles
\begin{verbatim}
have eA (...) : lhsA = rhsA := by exact h ...
have eB (...) : lhsB = rhsB := by
  have lifted := congrArg C (eA ...)
  exact lifted.trans ...
...
exact finalEquality
\end{verbatim}
where actual variable binders, contexts, and equation sides are printed from
the replay graph.  No saturation data structure or KBO comparison is imported
into \lean.  The certificate states only the equational consequences needed
for the final chain.

This proof shape also supports robust recovery.  If an exact term is difficult
for elaboration because an implicit argument is ambiguous, the robust printer
adds local type information or a small closing alternative around that step.
It never replaces a missing inference with an axiom.  Because reachable nodes
are emitted once, a clause used by several descendants is shared by name rather
than duplicating its derivation.

\subsection{Acceptance as the commit point}

In Solo mode the solver sends the complete source in a judge request.  The
proxy writes the judge-controlled problem module, compiles the candidate, runs
dependency inspection, and returns one of the documented statuses.  Only the
accepted status commits a solution row.  All other statuses preserve enough
error text for robust re-emission or later fallback, subject to the remaining
wall-clock budget.

Marathon cannot make interactive judge calls.  Its commit point is therefore
split: the solver durably appends a complete candidate row, and the runner
checks the last complete candidate for each identifier after process exit.
The append-and-synchronize protocol ensures that termination between problems
cannot corrupt already emitted source.  This difference changes scheduling
but not trust: the same verifier and proof policy decide both tracks.

\section{Evaluation}
\label{sec:evaluation}

\subsection{Method and provenance discipline}

Unless marked hosted, measurements in this section are local runs of the
official runner and deterministic \lean judge.  The main regression used
official repository revision \texttt{2848228ff490...}; the harness completed
without errors on the operator host.  The tested solver is v2.5, SHA-256
\texttt{f2392533c9f4...}, with byte count given in the abstract.  Each result
file is bound by the repository's \code{PROVENANCE.json}.  The short hashes in
the tables are prefixes of the full SHA-256 values in that file.

The reported wall-clock time is the sum of per-row runner measurements, not
elapsed parallel makespan and not a hardware-independent cost model.  Solver stages
with deadlines may take different paths under heavier load.  ``LLM'' counts
language-model calls attributed by the runner; a zero means the deterministic
cascade produced the accepted certificate.

\subsection{Six public sets}

Table~\ref{tab:public} reports the final six-set regression.  All rows were
accepted, every accepted row had one judge call, and no row used the fallback
model.  The sum of the set sizes is 1,889.  The earlier frozen v1 solver
accepted 1,875 of those rows; v2 adds the structured false stages and G3 proof
engine described above.  We report v1 only as development context, not as a
comparison against another entrant.

\begin{table*}[t]
\caption{Local official-runner regression for solver v2.5.  ``Wall'' is the
sum of per-row wall-clock seconds.  Each row names its immutable ledger and
SHA-256 prefix.}
\label{tab:public}
\small
\setlength{\tabcolsep}{4pt}
\begin{tabularx}{\textwidth}{@{}lrrrrY@{}}
\toprule
Set & Accepted & LLM & Judge & Wall (s) & Ledger provenance \\
\midrule
\code{sample\_20}  & 20/20 & 0 & 20 & 67.53 & \prov{results/v2\_sample\_20\_official\_2848228.json}{b17eaff452d2...} \\
\code{sample\_200}  & 200/200 & 0 & 200 & 964.12 & \prov{results/v2\_sample\_200\_official\_2848228.json}{dcf830de6199...} \\
\code{hard1}  & 69/69 & 0 & 69 & 559.81 & \prov{results/v2\_hard1\_official\_2848228.json}{7724566d1ceb...} \\
\code{hard2}  & 200/200 & 0 & 200 & 1250.67 & \prov{results/v2\_hard2\_official\_2848228.json}{3e68548945d4...} \\
\code{hard3}  & 400/400 & 0 & 400 & 2857.38 & \prov{results/v2\_hard3\_official\_2848228.json}{c9f98750e018...} \\
\code{normal}  & 1000/1000 & 0 & 1000 & 4128.3 & \prov{results/v2\_normal\_official\_2848228.json}{27c756482d15...} \\
\bottomrule
\end{tabularx}
\end{table*}

The totals establish regression coverage of these public artifacts only.  The
private evaluation set is separate, and public-set tuning is possible.  We do
not treat Table~\ref{tab:public} as a hidden-set estimate.

\subsection{Component-level interpretation}

The superseded v1 baseline left fourteen public residuals: eleven true
implications without a deterministic proof inside its budgets and three false
implications without a model in its available families and carriers.  The v2
development was organized around those failure classes.  G3 plus the
tautology-deletion correction supplies certificates for the true residuals;
the large-carrier, Latin-propagating, structured, and infinite-carrier paths
supply witnesses for the false residuals.  Replaying the residual set alone
records fourteen accepted certificates, whereas the six complete sets provide
a regression check that stage ordering and printers preserve the previously
accepted cases.

This history is diagnostic, not a controlled ablation study.  Several changes
were introduced together, and a row may now be solvable by more than one stage.
The paper therefore attributes mechanisms only where the engineering notes contain
a direct replay, such as \code{hard3\_0314} for tautology deletion and
\code{hard1\_0062} for Latin propagation.  It does not assign an aggregate
percentage gain to individual components.

The order of the cascade also affects the observed wall totals.  A true row
may incur the shallow false-family cost before receiving a quick proof; a false
row may incur direct proof checks before finite-model search.  These costs are
intentional because the early stages reduce resource use across both branches:
inexpensive structured countermodels avoid proof saturation, while direct proof
patterns avoid table search.  The six-set totals measure the integrated policy,
not isolated engine speed.

\subsection{Distribution drill, Marathon, and hosted playground}

The official-distribution drill uses the four published Stage 1 evaluation
splits that were named as Stage 2 scoring categories.  Each split contains
published ground truth and uses the asterisk input encoding.  Table
\ref{tab:drill} reports full local judge acceptance and full agreement with
those labels.  Stage 2 does not reuse these problems, so the drill measures
format and distribution readiness rather than evaluation transfer.

The canonical Marathon measurement used the official batch runner and its
\code{normal\_100} manifest.  All rows were accepted without tokens.  The
hosted-playground row is different in kind: it was run through the official
playground interface on the hosted judge and is recorded as an external
operator attestation, because the playground emits no repository-committable
ledger.  Across the four evaluation categories the attested runs record
acceptance of every attempted problem with no rejected or errored rows and no
language-model calls.  It is explicitly not an official leaderboard score.

\begin{table*}[t]
\caption{Additional measurements.  Local drill and Marathon rows name their
ledger files and SHA-256 prefixes.  The hosted row is an operator
attestation recorded in the \code{hosted\_playground\_measurement} entry of
\code{PROVENANCE.json}; the playground interface did not emit a repository
ledger file.}
\label{tab:drill}
\small
\setlength{\tabcolsep}{4pt}
\begin{tabularx}{\textwidth}{@{}lrrrY@{}}
\toprule
Measurement & Accepted & LLM/tokens & Wall statistic & Ledger provenance \\
\midrule
\code{evaluation\_order5} & 200/200 & 0 calls & sum 583.29 s & \prov{results/official\_distribution\_drill/evaluation\_order5\_results.json}{422ecdf5a856...} \\
\code{evaluation\_extra\_hard} & 200/200 & 0 calls & sum 1735.26 s & \prov{results/official\_distribution\_drill/evaluation\_extra\_hard\_results.json}{99d16aecd72f...} \\
\code{evaluation\_hard} & 200/200 & 0 calls & sum 889.03 s & \prov{results/official\_distribution\_drill/evaluation\_hard\_results.json}{4505248a3531...} \\
\code{evaluation\_normal} & 200/200 & 0 calls & sum 2288.06 s & \prov{results/official\_distribution\_drill/evaluation\_normal\_results.json}{2377747990d8...} \\
Marathon \code{normal\_100} & 100/100 & 0 tokens & full default budget & \prov{results/marathon/normal\_100\_summary.json}{28fc79269ede...} \\
Hosted playground (\code{evaluation\_normal}) & 200/200 & 0 calls & mean 4.89 s/problem & \prov{PROVENANCE.json: hosted\_playground\_measurement}{artifact 7916cbc2...} \\
\bottomrule
\end{tabularx}
\end{table*}

The hosted record further reports means by verdict: 2.69 seconds for
false rows and 7.08 seconds for true rows, with maxima 6.5 and 23.8 seconds,
respectively.  Those values include hosted orchestration and certificate
checking as exposed by the playground, and should not be compared directly to
the local prover-only benchmark below.

\subsection{Focused prover measurement}

To test the G3 search independently of the full cascade, development used a
corpus of 254 ETP pairs selected because an existing Vampire derivation was
available and the earlier prover found them difficult.  Under a thirty-second
per-pair prover cap, the optimized engine found all 254 derivations; maximum
prover time was 2.7 seconds and the sum was sixteen seconds.  On all 819 public
true problems under the quick cap, the prover found all proofs with median one
millisecond and maximum 1.2 seconds.  These are prover-only measurements on the
development machine, not judge wall-clock time and not a comparison claiming
superiority over Vampire.  Vampire served as a source of hard instances and
derivation guidance.

These focused results are recorded in
\code{docs/\allowbreak TRUE\_\allowbreak SIDE\_\allowbreak G3\_\allowbreak PROVER.md}; they have no standalone hash-bound
runner ledger and are therefore not included in the ledger tables.

\subsection{Bounded language-model measurement}

Before the v2 deterministic extensions, a fixed-configuration run exercised
the organizer-pinned \code{gpt-oss-120b} fallback on six unsolved
\code{sample\_20} residuals.  It solved zero of the six.  Pilot and canonical
runs also produced different first attempts despite the advertised zero
temperature and fixed seed.  The canonical run cost \$0.1350 in total,
measured from provider credits-balance deltas rather than runner logs ---
about \$0.023 per residual item under the sixteen-round cap.  Per-call
latency and token telemetry are recorded as unavailable in the run notes, so
no per-call figures are reported here.  We report this bounded negative measurement
of one model, provider route, solver version, prompt, and residual set.  It
does not support a general statement that language
models are ineffective for theorem proving or for this competition.

The executed pre-fix solver is bound in \code{docs/\allowbreak LIVE\_\allowbreak RUN\_\allowbreak NOTES.md} by
SHA-256 prefix \texttt{22a0bc84\allowbreak 6288...}; the canonical row data are in
\code{results/\allowbreak sample\_\allowbreak 20\_\allowbreak live.json}.  Because no ledger-file hash for this
historical run appears in the claims provenance, it is reported in prose rather
than the hash-bound tables.

\section{Honest Boundary and Reproducibility}
\label{sec:repro}

\subsection{Hosted result: \pending}

At the time of this draft there is no hosted leaderboard score or rank.  The
hosted playground is a diagnostic service and is labeled as such.  The six-set
and distribution-drill results are local official-runner measurements.  None
of them establishes transfer to the hidden evaluation distribution.

There is also no completeness theorem for either branch.  Structured
countermodels cover chosen families, the finite-model search has carrier and
time bounds, and the superposition implementation deepens only until its
deadline.  Deep-stage outcomes can vary with machine load even though the
judge's verdict on a fixed certificate is deterministic.  We make no
performance-superiority claim over another solver or prover.  The empirical
statement is exactly that the archived certificates were accepted in the
recorded runs.

\subsection{Threats to validity}

The public suites influenced solver development.  The fourteen v1 residuals
motivated particular prover and countermodel changes, and the E168 drill
motivated the fixed central groupoid.  Full regression shows that those
changes integrate without losing earlier rows; it does not measure performance
on an independently sampled corpus.  The distribution drill is broader but is
still published Stage 1 data, and its ground truth may share construction
patterns with the public suites.

Wall-clock totals combine Python search, process overhead, and \lean checking
under one operator-host environment.  They are appropriate for detecting large
regressions within that setup but should not be read as portable runtimes.
Parallel worker load affects time-budgeted deepening, and filesystem caching
affects \lean startup.  The hosted-playground mean times partially address this
environment gap, but playground selection and daily interaction are not the
same protocol as final evaluation.

Finally, kernel checking validates each emitted theorem relative to the
judge's imported environment and allowed axioms.  It does not validate that
the benchmark labels, competition problem generator, or provenance narrative
are correct.  We mitigate the last concern with hashes and exact commands;
benchmark construction and judge implementation remain external assumptions.

\subsection{Hash-bound evidence}

\code{PROVENANCE.json} binds the frozen solver's SHA-256 and byte count, the
official repository and configuration revisions, each result-ledger hash, the
Marathon outputs, the compatibility corpus, and the hosted-playground record.
The human-readable claims ledger states the maximum set of supported claims
and separately lists prohibited inferences.  The repository script
\code{scripts/check\_freeze.py} recomputes the solver and ledger hashes, checks
accepted-row metadata, verifies the one-file submission layout, and fails
closed on drift.

This structure is intended to prevent stale measurements from being attributed
to a revised solver.  Historical v1 and intermediate v2
artifacts remain named as superseded.  The tables in this paper use the current
v2.4 bindings from \allowbreak\code{PROVENANCE.json}, even where an older narrative report
contains timings from a previous rerun.

\subsection{Replay procedure}

Reproduction requires two checkouts: this artifact and the official Stage 2
judge at the pinned revision.  In the artifact checkout, first run
\begin{verbatim}
python3 scripts/check_freeze.py
\end{verbatim}
to check immutable bindings.  In the official checkout, run
\code{bash scripts/setup.sh}, source \code{.env.judge}, and invoke the Solo
runner with a submission-only directory containing the frozen
\code{solver.py}.  For example, the public normal set uses
\begin{verbatim}
python3 -m pipeline.runner \
  --submission /path/to/submission \
  --problems examples/problems/normal.jsonl \
  --output /tmp/normal-results.json
\end{verbatim}
The Marathon replay uses \code{scripts/run\_marathon.py}, the canonical
\code{normal\_100.jsonl} manifest, and an output directory outside a read-only
checkout.  Recomputed output hashes should be recorded as a new run rather
than overwriting the archived ledgers.

The \lean 4.32 compatibility corpus was run in an external directory and is
described in the provenance record; it is not presently packaged as a
one-command repository test.  This is a reproducibility limitation.  The
official harness, freeze checker, ledger files, and public-runner commands are
available locally; hosted playground interaction and any future leaderboard
submission require organizer infrastructure.

\paragraph{A measured outer boundary.}
Shortly before submission the organizers released two additional official sets.
On \code{stage2\_stress\_test}, announced as mirroring the final leaderboard
configuration, the frozen solver accepts 200 of 200 problems with no
language-model calls; the set publishes no per-item reference labels, but the
per-category verdict counts match the announced composition of twenty-five
true and twenty-five false in all four categories, and every accepted verdict
is kernel-checked by the judge.  On
\code{research\_order5\_hard} --- one hundred research-tier problems in which
at least one side is an order-five Austin law, whose false direction provably
admits no finite countermodel, and whose ground truth is in part unknown even
to the organizers --- the frozen solver certifies none within its budgets.
Both ledgers are committed and hash-bound.  The pair delimits the system
precisely: complete coverage of the announced evaluation distribution, and a
research frontier, coinciding with open territory in the Equational Theories
Project, that finite methods do not reach.

\section{Related Work}

\paragraph{Equational theories.}
The Equational Theories Project builds a collaborative implication graph for
single magma identities and supplies formal proofs, finite models, and curated
problem data~\cite{etp}.  The SAIR task derives its laws and implication
setting from this project but imposes a distinct online solver and certificate
interface.  Cazares analyzes Stage 1 of the challenge and its prediction
setting~\cite{cazaresStage1}.  Stage 2 changes the observable output from a
label to a judge-accepted proof or countermodel.

Knuth's study of central groupoids is directly relevant to the E168 witness
family~\cite{knuthCentral}.  Our use is narrower than the structural theory:
we embed one checked non-natural finite model because natural constructions did
not separate the goals encountered in the drill.  The infinite-carrier Austin
models similarly draw on constructions represented in the Equational Theories
Project rather than proposing a general finite-model theorem.

{\sloppy
\paragraph{Superposition provers.}
Vampire~\cite{vampire}, E~\cite{eprover}, and Prover9~\cite{prover9} are
mature first-order theorem provers implementing
variants of resolution, superposition, completion, simplification, and
given-clause saturation~\cite{vampire,eprover,prover9}.  Our engine borrows
standard ideas---\allowbreak ordered inference, KBO, demodulation, discrimination-\allowbreak tree
retrieval, and age/weight selection---but specializes them to unit equations
over one binary symbol.  The system-level difference is the output contract:
the search trace is compiled into a positive \lean equality proof acceptable
under the competition policy.  We do not compare throughput or coverage
against these systems.
\par}

\paragraph{Proof assistants and certificate checking.}
\lean is both the certificate language and the final trusted checker
\cite{lean4}; Mathlib supplies the surrounding library in the official judge
environment~\cite{mathlib}.  The solver follows the proof-producing automation
pattern in which a fast untrusted procedure returns an object reconstructed or
checked by a small trusted kernel.  Finite countermodels are likewise proof
objects: exhaustive reflection closes the concrete finite statement, while
infinite witnesses use explicit lemmas.

\paragraph{Distillation artifacts.}
The competition is motivated by work on distilling many-shot context into a
compact cheat sheet~\cite{hondaCheatSheet}.  The artifact here is executable
rather than a natural-language sheet: algebraic families, search control, and
proof printers are distilled into one source file.  The fixed-configuration
language-model measurement is included to document the development path, not
to argue that symbolic search and language models are mutually exclusive.

\section{Conclusion}

The SAIR EQT2 Stage 2 contract rewards an answer only when the same artifact
also supplies a machine-checked reason.  The solver described here meets that
contract with a cheapest-first combination of structured and irregular
countermodels, infinite witnesses, and ordered unit superposition, followed by
replay through the \lean kernel.  Hash-bound local regressions account for all
measured public and distribution-drill rows, and a separately labeled hosted
playground measurement exercises the announced toolchain.  The remaining
boundary is explicit: no leaderboard result, hidden-set claim, completeness
theorem, or comparative claim is available at this draft stage.

\ifplainarticle
  \bibliographystyle{plainnat}
\else
  \bibliographystyle{ACM-Reference-Format}
\fi
\ifanonsubmission\else
\section*{Author contributions (CRediT)}
Haobo Ma: conceptualization, methodology, software (solver architecture and the
systems foundation), investigation, validation.
Wenlin Zhang: conceptualization, methodology, software, investigation,
validation, data curation, project administration, writing (original draft).
Manuel Israel C\'azares: validation (solver freeze verification --- hash, file
size, and single-file contract compliance; evidence-ledger auditing across the
reviewed freezes); methodology (review of experimental design and the
claims--evidence boundary); writing (review and editing --- full manuscript
review and critical review of the presentation and claims).
\fi

\bibliography{references}

\end{document}